\pdfoutput=1

\documentclass[11pt]{article}

\usepackage[]{acl}

\usepackage{times}
\usepackage{latexsym}

\usepackage[T1]{fontenc}

\usepackage[utf8]{inputenc}

\usepackage{microtype}
\usepackage{graphicx}
\usepackage{algorithm}
\usepackage{algpseudocode}
\usepackage{booktabs}
\usepackage{multirow}
\usepackage{arydshln}
\usepackage{amsmath}
\usepackage{hyperref}
\usepackage{enumitem}
\usepackage{amssymb}
\usepackage{pifont}
\newcommand{\cmark}{\ding{51}}%
\newcommand{\xmark}{\ding{55}}%
\usepackage{xcolor}

\newcommand\compact{\textsc{Pearl}}
\usepackage{tablefootnote}
\definecolor{purplecolor}{RGB}{102, 0, 102}
\definecolor{graycolor}{RGB}{153, 153, 153}

\title{\compact: Prompting Large Language Models to \\ Plan and Execute Actions Over Long Documents}

\author{
  Simeng Sun$^1$\thanks{\ \ Work partially done during an internship at Microsoft.}\hspace{3mm} Yang Liu$^2$\hspace{3mm} Shuohang Wang$^2$ \hspace{3mm} \textbf{Chenguang Zhu}$^2$\hspace{3mm} \textbf{Mohit Iyyer}$^1$\\
  University of Massachusetts Amherst$^1$ \hspace{1em} Microsoft Research$^2$ \hspace{1em} \\
  \texttt{\{simengsun, miyyer\}@umass.edu} \\
  \texttt{\{yaliu10,shuohang.wang,chezhu\}@microsoft.com} \\
  }

\begin{document}
\maketitle

\newcommand\compactH{P1}
\newcommand\compactM{P2}
\newcommand\compactMNR{A1}
\newcommand\compacttwty{A2}
\newcommand\compactoneforty{A3}
\newcommand\compactexec{A4}
\newcommand\compactprependplan{A5}

\newcommand{\sscomment}[1]{\textcolor{cyan}{\bf \small [ #1 --SS]}}
\newcommand{\micomment}[1]{\textcolor{red}{\bf \small [ #1 --MI]}}
\newcommand{\cz}[1]{\textcolor{blue}{\bf \small #1 --CZ}}
\newcommand{\yang}[1]{\textcolor{red}{\bf \small Y: #1}}

\begin{abstract}
Strategies such as chain-of-thought prompting improve the performance of large language models (LLMs) on complex reasoning tasks by decomposing input examples into intermediate steps. However, it remains unclear how to apply such methods to reason over \emph{long input documents}, in which both the decomposition and the output of each intermediate step are non-trivial to obtain. In this work, we propose \compact, a prompting framework to improve reasoning over long documents, which consists of three stages: action mining, plan formulation, and plan execution. More specifically, given a question about a long document, \compact\ decomposes the question into a sequence of actions (e.g., \texttt{\footnotesize SUMMARIZE}, \texttt{\footnotesize FIND\_EVENT}, \texttt{\footnotesize FIND\_RELATION}) and then executes them over the document to obtain the answer. Each stage of \compact\ is implemented via zero-shot or few-shot prompting of LLMs (in our work, GPT-4) with minimal human input. We evaluate \compact\ on a challenging subset of the QuALITY dataset, which contains questions that require complex reasoning over long narrative texts. \compact\ outperforms zero-shot and chain-of-thought prompting on this dataset, and ablation experiments show that each stage of \compact\ is critical to its performance. Overall, \compact\ is a first step towards leveraging  LLMs to reason over long documents.\footnote{\url{https://github.com/SimengSun/pearl}}

\end{abstract}
\section{Introduction}

Performing complex reasoning over long input documents often requires forming high-level abstractions of the text (e.g., plots and themes in a narrative) and then conducting a variety of inferences on top of those abstractions~\citep{graesser1994constructing}. Consider the following question about the story ``Breakaway'' from the QuaLITY dataset~\citep{pang-etal-2022-quality}:
\begin{quote}
\footnotesize
What part of the final scene best connects to the story's opening conversation?
\end{quote}
To answer this question, we need to gather, evaluate, and synthesize information from across the story, which motivates decomposing the question into a \emph{plan of actions}, as in:
\begin{quote}
\footnotesize
\begin{enumerate}[leftmargin=*]
\item Identify all participants in initial conversation.
\item Summarize the initial conversation.
\item Summarize events and themes of final scene.
\item Summarize roles of conversation participants in final scene.
\item Identify and rank connections between conversation and final scene.
\end{enumerate}
\end{quote}

\begin{figure}[!t]
    \centering
    \includegraphics[width=\linewidth]{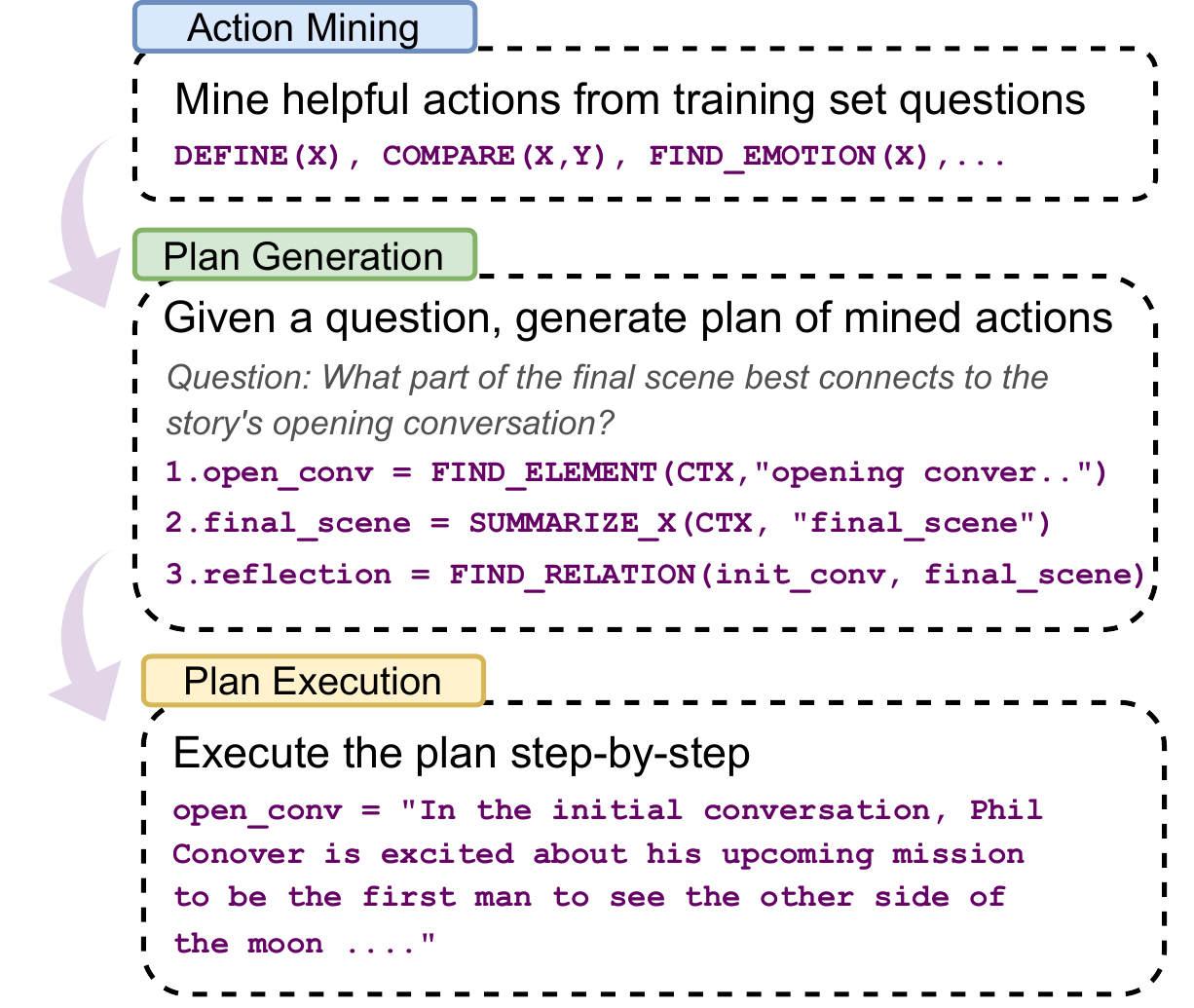}
    \caption{High-level overview of our framework \compact. Each stage in \compact\ is achieved via zero-shot or few-shot prompting of an LLM (in our work, GPT-4). We also provide \textcolor{purplecolor}{\texttt{ \small{example outputs}}} from each stage.}
    \label{fig:figure_1}
\end{figure}
\noindent Each action in the above plan varies in complexity, from simple lookup-style actions (Step 1) to more challenging query-focused summarization (Steps 2-4) and conceptual linking (Step 5) actions that require deep narrative understanding.

Given the rapidly advancing capabilities of large language models (LLMs), how can we use them to answer questions like these? While we could directly prompt LLMs to generate the answer, prior work on simpler reasoning-based tasks shows that this method is inferior to Chain-of-Thought prompting~\citep[][CoT]{wei2022chain}, which encourages the LLM to provide step-by-step explanations and intermediate outputs before producing the answer. Unfortunately, CoT is not well-suited for tasks involving complex reasoning over long input documents, as both the decomposition of the original question and the intermediate outputs of each step are non-trivial to obtain, as in the above example.  

Given the difficulty of obtaining plans and intermediate explanations for long documents, one potential solution is to delegate this task to smaller \emph{executable} modules instead of forcing the LLM to come up with all of them at once. In this work, we introduce \compact, a framework that combines \textbf{P}lanning and \textbf{E}xecutable \textbf{A}ctions for \textbf{R}easoning over \textbf{L}ong documents. Each stage of \compact\ --- action mining, plan decomposition, and plan execution --- is implemented by applying zero-shot or few-shot prompting to an LLM. The stages (Figure~\ref{fig:figure_1}) can concisely be described as follows:

\begin{enumerate}
\item \textbf{Action mining:} An LLM is prompted to come up with simple actions that can help solve questions from an input training dataset. Unlike predefined ``toolboxes'' in methods such as Toolformer~\citep{schick2023toolformer} or ReACT~\citep{yao2023react}, the action set in \compact\ is also generated by an LLM.
\item \textbf{Plan generation:} Given an input test question, an LLM generates an executable plan consisting of a series of actions selected from the action set produced in the previous stage. The plan is formatted as a simple program in which the execution result of one action can serve as an argument to future actions, which enables complex composition.
\item \textbf{Plan execution:} The LLM executes the plan action-by-action via a prompt template that includes an action and the long-form input document. Note that this is the only stage that includes the document, as the other stages operate over just questions. 
\end{enumerate}

\begin{table*}[]
    \centering
    \scalebox{1}{
    \begin{tabular}{lccccc}
    \toprule
       Prompting Methods  & \begin{tabular}[c]{@{}c@{}}Explicit\\plan\end{tabular} & \begin{tabular}[c]{@{}c@{}}
            Iterative\\
            prompting
       \end{tabular}   & \begin{tabular}[c]{@{}c@{}}Does not rely on \\external tools\end{tabular} & \begin{tabular}[c]{@{}c@{}}Long \\documents\end{tabular}  \\\midrule
       Chain-of-Thought~\citep{wei2022chain} & \xmark & \xmark & \cmark & \xmark\\
       Program-of-Thought~\citep{chen2022program} & \xmark & \xmark  & \xmark & \xmark\\
       Self-Ask~\citep{press2022measuring} & \xmark & \cmark & \xmark &  \xmark\\
       Toolformer~\citep{schick2023toolformer}  & \xmark &  \xmark   & \xmark & \xmark\\
       ReAct~\citep{yao2023react} & \xmark & \cmark & \xmark & \xmark \\
       Plan-and-Solve~\citep{wang2023planandsolve} & \cmark & \xmark & \cmark & \xmark\\
       \compact\ (\emph{this work}) & \cmark & \cmark & \cmark & \cmark \\
    \bottomrule
    \end{tabular}}
    \caption{Comparison of \compact\ to other recently-proposed prompting techniques. \compact\ is the only one designed for and evaluated on tasks that require complex reasoning over long documents. }
    \label{tab:related_comparison}
\end{table*}

We demonstrate \compact's effectiveness on a challenging subset of QuALITY~\citep{pang-etal-2022-quality}, a reading comprehension dataset that contains questions about long-form articles. While QuALITY is originally a multiple-choice dataset, we reformulate it into a generation task: given a question and an article, an LLM is asked to generate a free-form answer. As a proxy for measuring answer correctness, we adopt a similar approach to~\citet{wang-etal-2020-asking} by asking the LLM to map its generated answer to one of the multiple choice options, which allows us to compute its accuracy.

Prompting LLMs with \compact\ yields more accurate and comprehensive answers than those generated by directly prompting the LLM to answer the question, particularly for questions that require reasoning over the full long document. This result is particularly impressive given the potential for error propagation in the \compact\ framework: as each stage is implemented via an LLM, errors in plan formulation or execution can significantly affect the output answer. To further verify the integrity of the plans, we perform human evaluation by asking annotators to provide feedback and ratings; annotators generally find the plans to be reasonable, although a small percentage contain unnecessary actions or omit critical actions. Overall, we hope \compact\ further opens the door towards using LLMs for complex reasoning over long documents.

\section{Related work}

Our work builds on recent LLM prompting research and also connects to work on reasoning over long documents. Before describing \compact, we first survey related papers to contextualize our work within this fast-moving field.

\paragraph{Prompting methods:} Recently, the capabilities of large language models~\citep{gpt-3, zhang2022opt, touvron2023llama} have significantly increased as a result of learning from instructions or feedback~\citep{stiennon2022learning, ouyang2022training,chung2022scaling} to better align their outputs to human preferences. When provided with well-crafted prompts, such as chain-of-thought~\citep{wei2022chain} explanations, these state-of-the-art models exhibit impressive reasoning abilities. A plethora of new prompting techniques (Table~\ref{tab:related_comparison}) has been introduced lately to unlock more capabilities of LLMs via leveraging exteral tools~\citep{chen2022program,schick2023toolformer,lu2023chameleon}, problem decomposition~\citep{press2022measuring,dua-etal-2022-successive,khot2023decomposed,yao2023react}, self-reflection and self-refinement~\citep{huang2022large,shinn2023reflexion,madaan2023selfrefine}, planning~\citep{yao2023tree,wang2023planandsolve,long2023large}, and other techniques~\citep{yoran2023answering,wang2023selfconsistency,zhou2023leasttomost}.

\paragraph{Reasoning over long documents:} Large language models have showcased remarkable reasoning capabilities~\citep{huang2022reasoning}, including mathematical reasoning~\citep{gsm8k}, commonsense reasoning~\citep{talmor-etal-2019-commonsenseqa}, and symbolic reasoning~\citep{nye2021improving}. Most of these tasks do not involve long context inputs, and thus they are able to benefit from few-shot in-context CoT prompting. In this paper, we primarily focus on tasks that contain long input contexts~\citep{kocisky-etal-2018-narrativeqa,dasigi-etal-2021-dataset,shaham-etal-2022-scrolls,sun-etal-2022-conditionalqa}, specifically generative question answering based on long input articles. To address the absence of reliable evaluation for long-form QA~\citep{krishna-etal-2021-hurdles}, ~\citet{stelmakh-etal-2022-asqa} proposes automatic metrics for evaluating the correctness of the answer, whereas in this work, we use LLM-based evaluation by taking advantage of the multiple-choice setup of existing QA dataset. Prior to the shift to prompting-based methods, approaches including contrastive learning-based sequence-level objectives ~\citep{caciularu-etal-2022-long}, iterative hierarchical attention~\citep{sun2021iterative}, and joint modeling of machine reading and answer generation~\citep{su-etal-2022-read} have been employed to enhance long-context question answering.

\section{\compact: Planning and Executing Actions for Reasoning over Long Documents} \label{sec:compact}

We are interested in using LLMs to solve tasks that require complex reasoning over long documents.\footnote{As there is no consensus on what is ``long'', we consider it to mean documents of several thousands of tokens in length.} In this paper, we focus on the task of answering questions about long-form narratives.
Most prompting strategies that aim to improve the reasoning abilities of LLMs (e.g., CoT) are not applicable to this task due to the length and complexity of the input document. In this section, we specify our \compact\ framework, which consists of three LLM-implemented stages that mine actions from a training corpus, formulate plans to answer held-out questions, and then execute the resulting plans to obtain answers. 

\subsection{Action mining}  \label{sec:action_mining}
\definecolor{c1}{RGB}{201, 230, 244}
\setlength{\fboxsep}{0pt}
\begin{figure}
    \centering
    \includegraphics[width=\linewidth]{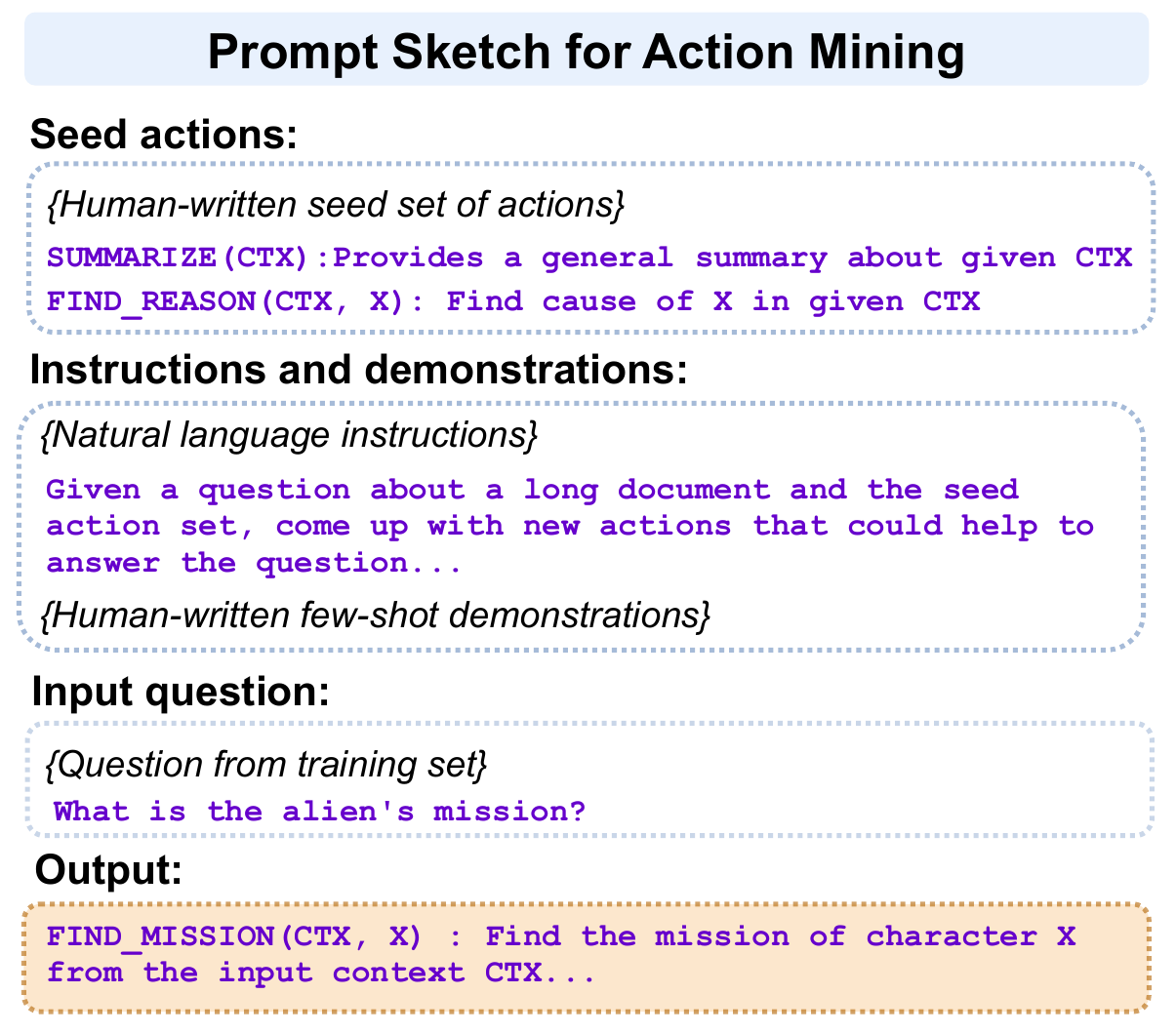}
    \caption{Prompt sketch for action mining. It comprises human-written seed actions set and instructions, as well as question for which LLM will extract action(s) from. Finally, we also present an example mined action. More details can be found in the Appendix~\ref{sec:prompts}.}
    \label{fig:action_mining}
\end{figure}
\begin{figure}[!t]
    \centering
    \includegraphics[width=\linewidth]{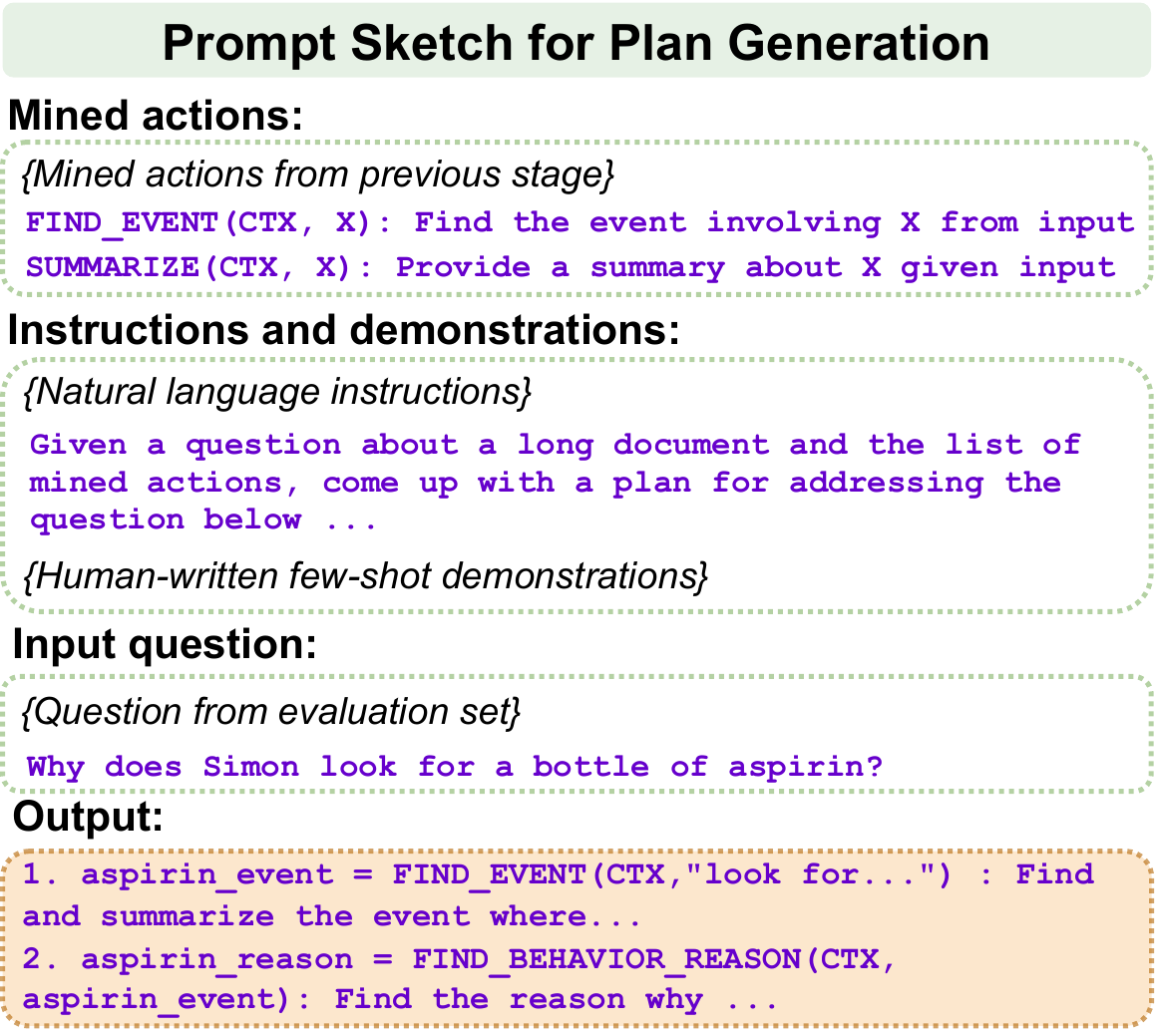}
    \caption{Prompt sketch for plan generation. In the prompt, we include the list of actions mined from previous stage in-context, natural language detailing the task, and few-shot examples guiding the plan generation.}
    \label{fig:plan_generation}
\end{figure}
In many prior prompting techniques such as ReACT and Toolformer, the LLM is able to query external APIs (e.g., Wikipedia search or a calculator) to solve a given task. Unlike these works, which assume a predefined action space, \compact\ mines actions directly from data of similar distribution (in our case, training set questions of QuALITY). As shown by prior research~\citep{graesser1994constructing}, answering complex queries over long documents requires specific reasoning techniques; as further evidence,~\citet{xu-etal-2022-answer} demonstrates the presence of various discourse structures in good answers to long-form questions on Reddit. Learning dataset-specific actions enables \compact\ to scale to different domains and tasks, as user queries may differ considerably in terms of complexity. Moreover, mining actions from training set can reduce human efforts in designing new actions. In this work, we define an ``action'' as a basic unit for long document reasoning. To obtain these actions, we first manually create a small set of \emph{seed} actions to use as demonstrations.\footnote{See prompt for QuALITY action mining in Appendix~\ref{sec:prompts}} Then, as shown in Figure~\ref{fig:action_mining}, given an example question, we feed it along with the seed actions and instructions to the LLM to generate more task-specific actions. Each \texttt{\footnotesize ACTION} is formatted as a programmatic function with input arguments and is followed by a \textit{model-generated function definition in natural language}.
Below is an example action generated by the LLM:
\begin{quote}
    \footnotesize
    \texttt{\small ANALYZE(CTX, X, Y)} \#  \textit{Analyze the relationship, attitude, or feelings between X and Y given the input context CTX}
\end{quote}
\noindent After a full pass over example questions in the training data, we obtain a final set of actions and their corresponding definitions which are then incorporated into the prompt of the next stage.

\subsection{Plan generation} 
A plan serves as the guiding framework or outline for answering complex questions that may involve multi-step reasoning and/or global understanding of long documents. Given a question, as shown in Figure~\ref{fig:plan_generation}, we prompt an LLM to generate a plan based on the previously-mined action set. Each step of the plan is formatted as 
\begin{quote}
\footnotesize
    \texttt{output = ACTION(arg$_1$, arg$_2$, $\dots$)},
\end{quote}

\noindent where the \texttt{\small output} variable stores the result of the current \texttt{\small ACTION}
, and the \texttt{\small arguments} can be (1) the input document, (2) a string, or (3) an output variable from previous steps of the plan.
When generating the plan, we do not show the LLM the entire document as input, which provides ample space for incorporating few-shot in-context examples. 
Similar to the seed actions in the previous stage, we provide a small seed set of plans and allow the model to generate more demonstrations automatically. We provide more details in Section~\ref{sec:exp} about controlling the quality of model-generated in-context demonstrations.

\begin{figure}[!t]
    \centering
    \includegraphics[width=\linewidth]{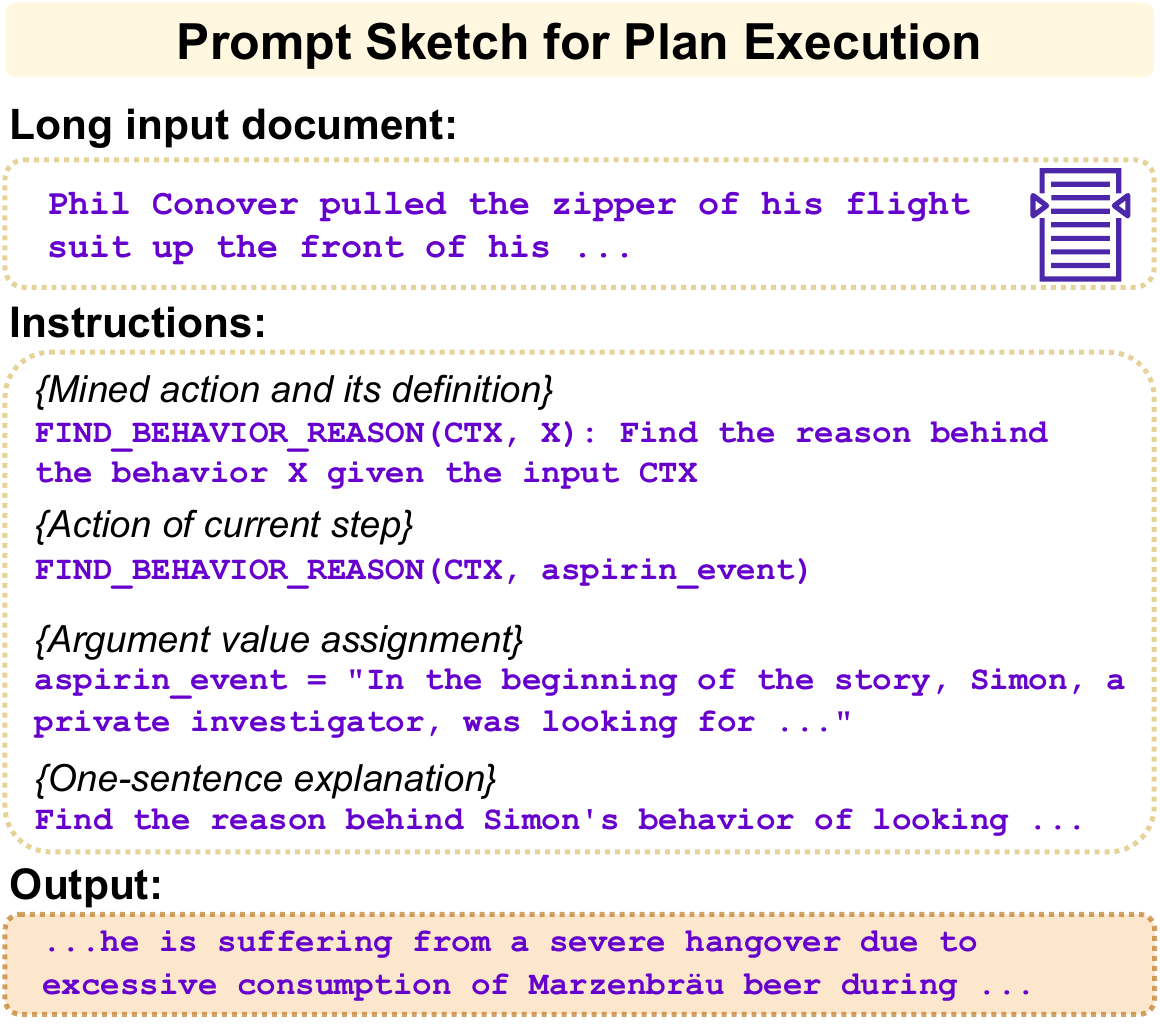}
    \caption{Prompt sketch for plan execution. This prompt contains multiple \textit{\{placeholders\}} that will be filled with output from previous stages. }
    \label{fig:plan_execution}
\end{figure}

\subsection{Plan execution} 
In the previous stage, the LLM generates a plan that serves as a blueprint for producing a response. To execute each step in the plan, we prompt the LLM with a template filled with output from previous stages. Concretely, as shown in Figure~\ref{fig:plan_execution}, to execute the action \texttt{\small FIND\_BEHAVIOR\_REASON}, the model fills in the prompt template with (1) the planned action and definition, (2) current action with specific input argument (e.g., \texttt{\small aspirin\_event})
, (3) assignment of argument name with output from previous stage (e.g., \texttt{\small aspirin\_event} = \texttt{\small ``in the beginning of the story, ...''}), and (4) a one-sentence instruction for the current step, all of which are generated by LLM. As the long input article is involved during this stage, the prompt is executed in a zero-shot manner. 

\subsection{Self-correction and self-refinement} Since the plans are generated by an LLM, they may be incorrectly formatted or of otherwise low quality. To address this issue, similar to~\citet{shinn2023reflexion}, we include a self-correction step prior to plan execution and a self-refinement step before incorporating model-generated plans as in-context few-shot examples. We implement a plan parser that returns relevant error messages when the plan does not conform to the defined format. The invalid plan as well as the error message are then passed into the LLM for correcting the plan's grammar.
To ensure the quality of model-generated in-context examples, we validate them by executing the plan and evaluating the generated answer with a task-specific scoring function (more details in Section~\ref{sec:exp_setup}). If the answer is rejected by the evaluation in the end, we pass the plan to LLM for further self-refinement before being included in the context as few-shot examples.
\section{Experiments}
\label{sec:exp}
\begin{table*}[]
    \centering
    \scalebox{1}{
\begin{tabular}{@{}lcccc@{}}
\toprule
        &  \begin{tabular}[c]{@{}c@{}}
             \textbf{\small \textsc{QuALITY}} \\
             \textbf{\small \textsc{Long}}
        \end{tabular}
        & \begin{tabular}[c]{@{}c@{}}
             \textbf{\small \textsc{QuALITY}} \\
             \textbf{\small \textsc{Short}}
        \end{tabular}& \textbf{\textsc{All}} & $p$-\textbf{val} \\  \midrule
\textbf{\textsc{Prompting Methods}}&      &       &       &       \\
\hspace{1em}GPT-4 zero-shot      & 64.3 & \textbf{79.1}  & 68.8  & -     \\
\hspace{1em}GPT-3.5 zero-shot (text-davinci-003)           & 45.5 & 56.3  & 48.8  & 0.000 \\
\hspace{1em}GPT-4 zero-shot chain-of-thought & 65.9 & 77.2 & 69.3 & 0.766\\
\hspace{1em}GPT-4 \compact\         & \textbf{70.9} & 77.8  & \textbf{73.0}    & 0.005 \\ \midrule
\textbf{Ablations of GPT-4 \compact} &      &       &       &       \\ 
\hspace{1em}w/o plan execution  & 67.3 & 77.2  & 70.3  & 0.295 \\
\hspace{1em}w/o self-refinement of plan demonstrations & 67.0   & 78.8  & 70.6  & 0.245 \\
\bottomrule
\end{tabular}
}
    \caption{We present baseline and \compact\ as well as ablation results on our generative subset of QuALITY questions. \textbf{Long} denotes the split where the questions require reasoning over long contexts to answer accurately. As we only evaluate on a subset, we also provide $p$-values to verify statistical significance against the zero-shot GPT-4 baseline.
    }
    \label{tab:main_res}
\end{table*}

We compare \compact\ to baseline methods (zero-shot answering and zero-shot CoT) on a challenging subset of the QuALITY Question-Answering dataset that requires reasoning over long articles of several thousands tokens. In this section, we describe our dataset selection, experimental setup, and model configurations.

\paragraph{Dataset selection:} We focus on the QuALITY QA dataset~\citep{pang-etal-2022-quality}, which is a multiple-choice QA task in the SCROLLS benchmark~\citep{shaham-etal-2022-scrolls}. 
However, to better simulate LLMs usage in real-world scenarios, we turn this dataset into a \emph{generative} task\footnote{We provide the performance of GPT-4 with standard multi-choice setup on the full QuALITY dev set in Appendix~\ref{sec:multi_choice_setup}.} in which an LLM does not have access to the choices and must instead generate a long-form answer. Then, we automatically map the generated answer back to one of the choices with an LLM to evaluate the accuracy as shown in Figure~\ref{fig:eval_gqa}.\footnote{In Appendix~\ref{sec:human_answer_map}, we confirm through human evaluation that GPT-4, the model we test, demonstrates considerable—but not perfect—agreement with human annotators for the answer mapping stage.} The accuracy of mapped answers serves as a proxy for assessing the correctness of the provided answer. 

QuALITY contains a diverse variety of questions, each of which is annotated with the amount of context from the document needed to answer the question. In contrast to questions that can be correctly answered with local context once a piece of information is located, as in
\begin{quote}
    \footnotesize
    Who found Retief and Magnan in the trees?
\end{quote}
\noindent we are more interested in questions that require reasoning over long context, as in:
\begin{quote}
    \footnotesize
    How would you describe the changes in tone throughout the passage?
\end{quote}
These questions constitute an interesting and difficult subset that, unlike more straightforward information seeking questions, require global understanding and reasoning over the document to provide accurate answers. Therefore, we select a subset of questions rated as requiring  long contexts to answer. In total, we create a dataset of 1K examples divided into two splits:\footnote{Human annotation score on the required context ranges from 1 to 4. Questions in the long split are those with average human annotation score $\geq 3$, questions in the short split have scores $<3$.} (1) \textbf{Long}: 330 examples from the dev set, 368 examples from training set, and (2) \textbf{Short}: 302 examples from dev set that do not require long contexts to answer; the latter forms a control dataset to make sure our methods  do not overly worsen performance on simpler questions.

\setlength{\fboxsep}{0pt}
\begin{figure}
    \centering
    \includegraphics[width=0.75\linewidth]{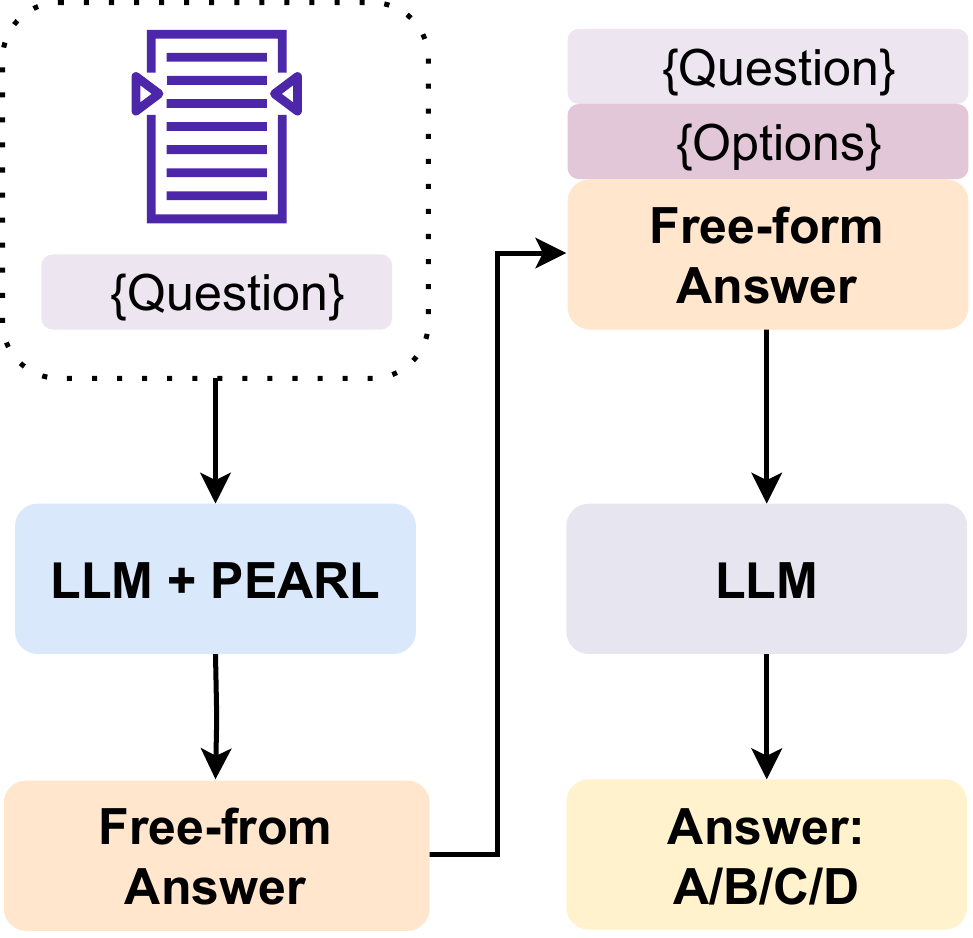}
    \caption{Generic illustration of our evaluation setup. Given the article and question, we prompt an LLM with \compact\ to generate a long-form answer, which is later mapped to one of QuALITY's multiple-choice options by the LLM itself.}
    \label{fig:eval_gqa}
\end{figure}

\subsection{Experimental setup} \label{sec:exp_setup}
As each of the stages in \compact\ has critical hyperparameters and implementation details, we describe our specific configurations here. 

\paragraph{Action mining:} We provide an LLM with seven seed actions and two in-context examples to demonstrate the required format for generating new actions.\footnote{We present the prompt template in Appendix~\ref{sec:prompts}} We collect new actions by passing all training set questions into the model, excluding those questions in our evaluation set. Ultimately, we obtain 407 actions and corresponding definitions, of which several are duplicates or overly specific, and in total exceeds GPT-4's maximum context window of 8K tokens. As such, we instruct the LLM to simplify and abstract over existing actions in order to reduce the total number of actions. After repeating this process twice,\footnote{After one round, the actions reduced to $\sim$140, and after four rounds to $\sim$20. We provide ablations on the number of actions in Section~\ref{sec:results}.} we reduce the number of actions to 81, which forms the final action set for \compact.

\paragraph{Self-correction retry limit:} Despite utilizing self-correction to validate the generated plan's syntax, it is still possible that the model fails to generate a plan in the correct format. In such cases, we force the model to revert to the zero-shot baseline approach. Out of 1K examples across various \compact\ variants, only 4 examples failed to parse within the retry count limit, which is within an acceptable range of failed examples.

\subsection{Baselines}
As existing sophisticated prompting methods require few-shot examples in-context, which is not feasible when long document is involved, we compare \compact\ with simple zero-shot baselines (GPT-4~\citep{openai2023gpt4} and GPT-3.5~\citep{ouyang2022training}), where we directly prompt the model to provide a detailed free-form answer. Additionally, we also evaluate zero-shot chain-of-thought prompting for GPT-4 by adding ``Let's think step-by-step,'' to the prompt.

\section{Main results}
\label{sec:results}

We discover that \compact\ significantly outperforms competing prompting methods on questions that require reasoning over long contexts, which demonstrates the utility of the planning module. We also observe a small drop in accuracy on questions that require only short contexts, possibly because the plans end up over-complicating what is a simple reasoning process. In this section, we dig deeper into the main results of our experiments, which are presented in Table~\ref{tab:main_res}. 

\begin{figure}[!t]
    \centering
    \includegraphics[width=\linewidth]{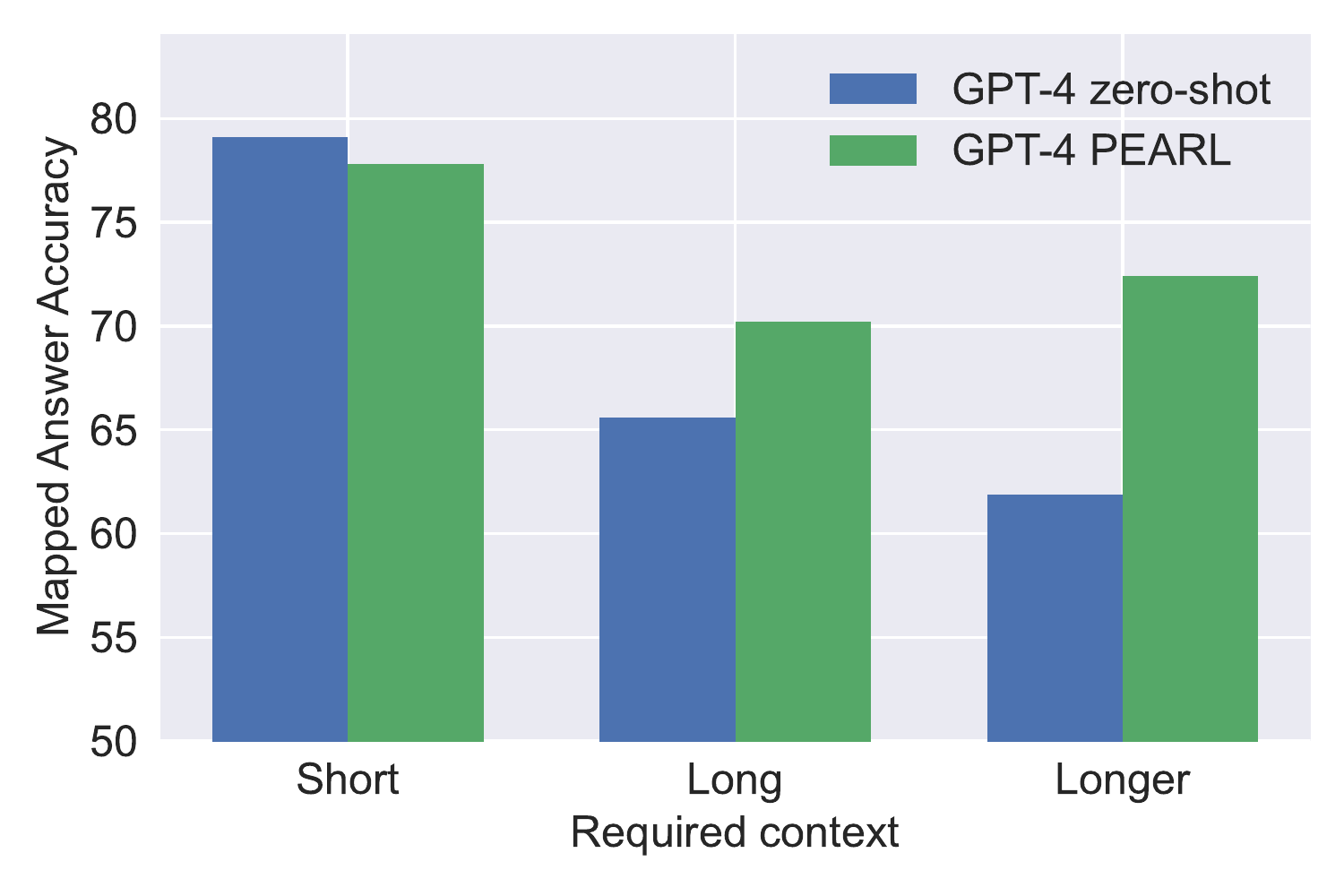}
    \caption{Accuracy by the amount of required context to answer,\footnotemark  as annotated by humans in QuALITY.}
    \label{fig:acc_by_ctxeval}
\end{figure}

\paragraph{\compact\ improves accuracy on long-document QA:}  Overall, \compact's accuracy is higher than that of all competing methods, particularly for the QuALITY split annotated by humans as requiring long contexts to answer (\textbf{Long}). Furthermore, we observe in Figure~\ref{fig:acc_by_ctxeval} that for questions marked by QuALITY workers as requiring the longest possible context, \compact\ improves substantially compared to the zero-shot GPT-4 baseline (72.4\% vs 61.9\%).
Our method's slightly diminished performance on the \textbf{short} split is likely due to both ``overthinking'' these simpler questions, as well as error propagation from plan execution steps as highlighted in Section~\ref{sec:analysis}. Finally, we point out that all methods achieve higher accuracies on the \textbf{Short} split compared to the \textbf{Long} split, indicating the challenging nature of this set of questions. 
\footnotetext{The short, long, and longer splits correspond to average annotation scores on the amount of required context [1, 3), [3, 3.5), and [3.5, 4), respectively.}

\begin{figure}[!t]
    \centering
    \includegraphics[width=0.9\linewidth]{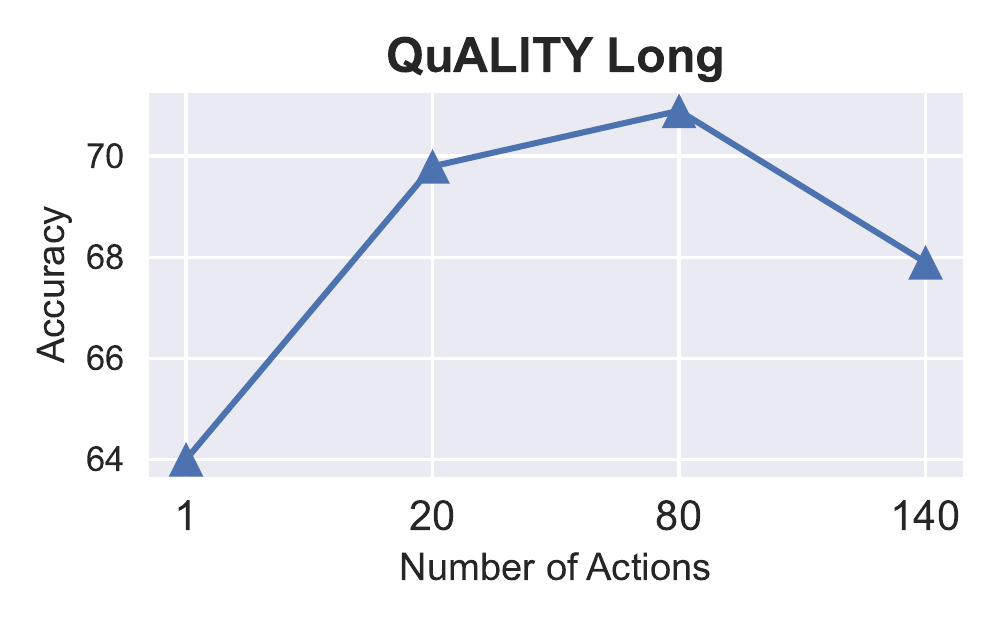}
    \caption{\compact\ accuracy given in-context action sets of various sizes. Having too few or too many actions impairs the performance.}
    \label{fig:num_action_ablat}
\end{figure}

\paragraph{Number of actions impacts performance:} In Figure~\ref{fig:num_action_ablat}, we show that the size of the action set is an important factor in \compact's performance. With just a single action (i.e., \texttt{\small EXECUTE} a free-form natural language instruction),\footnote{We additionally preserve the \texttt{\small CONCAT} action in this setting due to its necessity when aggregating execution results.} \compact's accuracy on the \textbf{Long} subset drops to 64\%. With too many actions (140 in the plot), its accuracy also degrades, likely because the action space is too fine-grained for the model to properly execute all actions. 
We note that the optimal number of actions likely differs from task to task, so it is an important hyperparameter to consider when tuning \compact.

\paragraph{Action execution is necessary:} Do we actually need to \emph{execute} the generated plans to answer these questions? Feeding just the generated plan to the model along with the question (minus any execution results) may still encourage the LLM to follow the plan's reasoning steps and generate a better answer. However, we observe that removing the execution results from the model's input reduces absolute accuracy by around 3 points, which suggests that it is important to perform multiple passes over the document to execute each action before answering the original question. With that said, we do observe a modest improvement over the GPT-4 zero-shot and CoT baselines ($\sim 2$ absolute points), which suggests that the plan itself is also valuable. 

\paragraph{Self-refinement improves performance:} To reduce human input, the majority of the plan generation demonstrations for \compact\ are generated by the LLM with self-refinement. We observe that self-refinement is critical to performance: without it, the overall accuracy drops nearly 3 absolute points (see ablations in Table~\ref{tab:main_res}), which highlights the importance of high-quality few-shot examples for plan generation. 
\section{Analysis} \label{sec:analysis}

In this section, we analyze the behavior of \compact\ by diving into the composition of its generated plans, its most preferred actions, and what types of questions it improves most on. We also offer a qualitative error analysis as well as a human evaluation on the correctness of the generated plans. 

\begin{table*}[]
    \centering
    \footnotesize
    \scalebox{0.9}{
    \begin{tabular}{p{0.15\linewidth} p{0.25\linewidth}p{0.3\linewidth}p{0.3\linewidth}}
\toprule
\textbf{Err. Category} & \textbf{Question} & \textbf{ Model Generated Plan or Answer} & \textbf{Explanation} \\
\midrule
\textbf{True Negative}  

 \textcolor{blue}{- Error in Plan}

(17.5\%)
& Does the tone of the passage shift at all, and if it does, how does it shift?

& \textcolor{purple}{(Plan)} ...
3. tone\_shift = COMPARE(CTX, tone\_initial, tone\_final, ``tone'') : Compare the initial and final tones of the passage to determine if there is a shift... \vspace{0.05em}& Since the plan only compares the initial and final tone, the final answer fails to capture the changes in between, thus leads to an incorrect answer. \\\hdashline

\vspace{0.05em}\textbf{True Negative} 

\textcolor{blue}{- Error in Exec.}

(55\%)
& \vspace{0.05em} How many adult characters have speaking roles? & \vspace{0.05em} \textcolor{purple}{(Answer)} In the input article, there are 3 adult characters with speaking roles...

 & \vspace{0.05em} 
 The correct answer involves two characters, whereas \compact's response mistakenly includes an additional name. The plan for this question is reasonable, but the problem stems from the execution of individual steps.
 
\vspace{0.05em}

\\\hdashline

\vspace{0.05em} \textbf{False Negative}

(12.5\%)& \vspace{0.05em}Does the story have a good ending? (Answer: Unclear, the story ends as Evelyn enters a dangerous situation) & \vspace{0.05em}\textcolor{purple}{(Answer)} ...However, the ending of the story is somewhat ambiguous and leaves several questions unanswered. For instance, it is unclear whether Evelyn will be able to successfully complete her mission 
... \vspace{0.05em} & \vspace{0.05em}In this example, the model output is correct, but is mapped to an incorrect distractor option, which contains direct contradictions with the model output.\\\hdashline

\vspace{0.05em}\textbf{Other} 

(15\%) & \vspace{0.05em} Who would most likely enjoy this story, of the following options? & \vspace{0.05em}\textcolor{purple}{(Answer)} The target audience of the input article is science fiction enthusiasts, particularly those who enjoy stories about space exploration, alien encounters... & \vspace{0.05em}The model output is not necessarily wrong in the absence of options. However, when provided with options during mapping stage, one of the other options is clearly better. \\

\bottomrule
\end{tabular}}
    \caption{Examples of errors exhibited by \compact\ answers.}
    \label{tab:err_examples}
\end{table*}

\begin{figure}
    \centering
    \includegraphics[width=\linewidth]{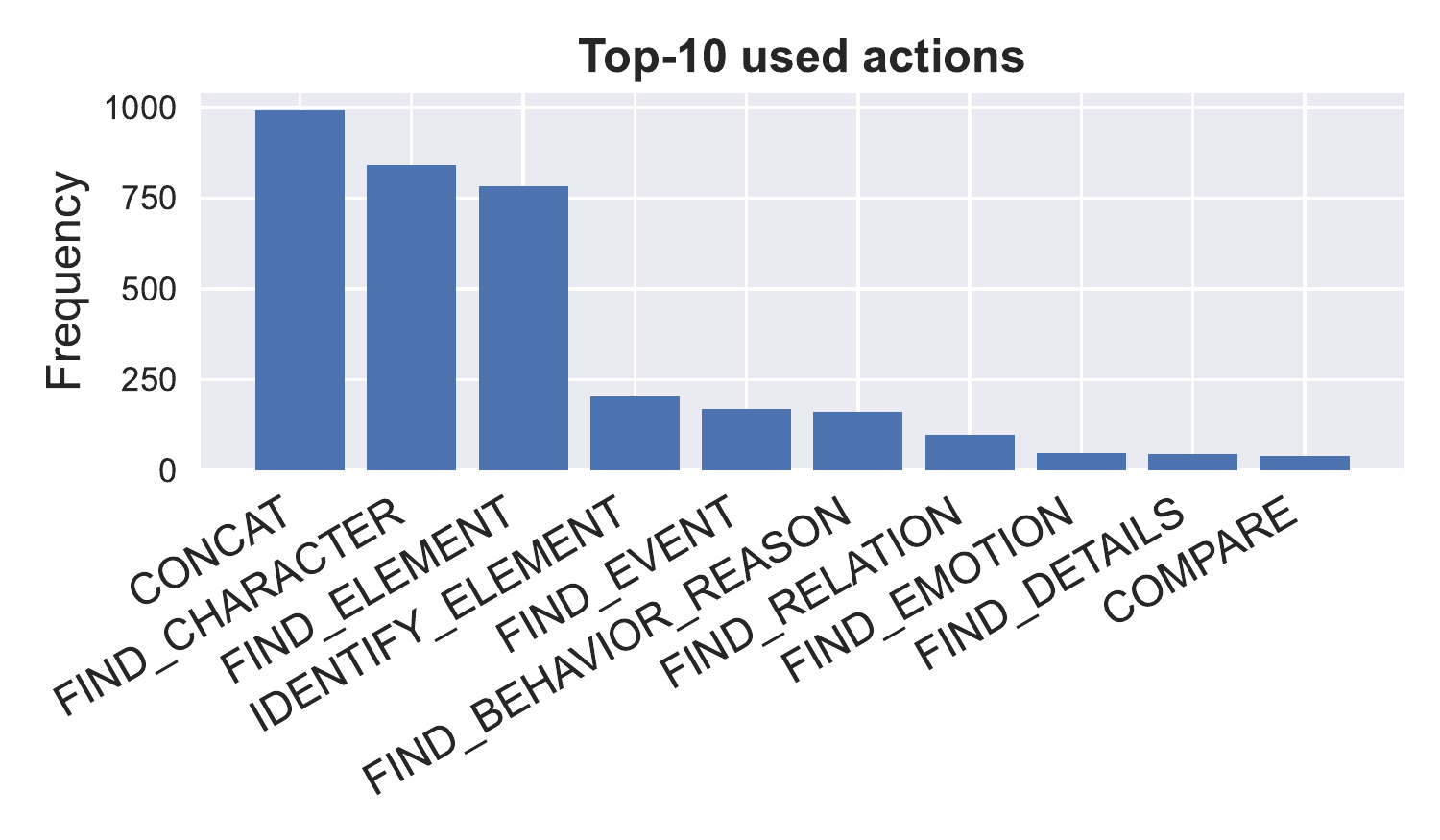}
    \caption{Top-10 most frequently used actions by \compact. }
    \label{fig:act_freq}
\end{figure}

\paragraph{Plan statistics:} Plans are roughly 4 actions long on average, with around 3.4 unique actions per plan. The most commonly used actions are shown in Figure~\ref{fig:act_freq}. Apart from the string concatenation action \texttt{CONCAT}, the most frequently used action is \texttt{FIND\_CHARACTER}, which can be convenient for understanding long literary text. Other less often used actions cover both those that can transfer across domains, e.g., \texttt{COMPARE}, and those specific to narrative understanding, e.g., \texttt{FIND\_EMOTION}.

\paragraph{Accuracy by reasoning types:} Since QuALITY questions require different reasoning strategies to solve, what types of reasoning does \compact\ help improve the most?  To this end, we further evaluate questions based on the type of reasoning required to answer them.\footnote{We prompt GPT-4 with the definition of each reasoning type presented in QuALITY's Appendix~\citep{pang-etal-2022-quality} and ask it to label each question with up to two reasoning types.} Table~\ref{tab:acc_by_reasoning} shows that \compact\ significantly improves three reasoning types: \emph{why} questions (reasoning about a cause), \emph{person} questions (reasoning about the person(s) involved in an event), and \emph{not/except} questions (e.g., ``which of the following is not a reason for...''). 
\begin{table}[!h]
    \centering
    \footnotesize
    \scalebox{0.95}{
    \begin{tabular}{@{}lrrr@{}}
\toprule
       & \textbf{Count}  & \begin{tabular}[c]{@{}c@{}}
            \textbf{GPT-4}  \\
            \textbf{\compact}
       \end{tabular} & \begin{tabular}[c]{@{}c@{}}
            \textbf{GPT-4}  \\
            \textbf{zero-shot}
       \end{tabular}  \\ \midrule
Description              & 320 & 0.73         & 0.73     \\
Why/reason               & 316 & 0.79$^*$         & 0.71$^*$     \\
Symbolism/interpretation & 262 & 0.73         & 0.70     \\
Person & 216 & 0.75$^*$         & 0.66$^*$     \\
Event  & 199 & 0.69         & 0.68     \\
Not/except               & 118 & 0.70$^*$         & 0.53$^*$     \\
How/method               & 100 & 0.74         & 0.73     \\
Relation                 & 89  & 0.71         & 0.65     \\
Entity & 74  & 0.64         & 0.68     \\
Numeric& 49  & 0.67         & 0.78     \\
Location                 & 32  & 0.59         & 0.59     \\
What if& 21  & 0.71         & 0.76     \\
Object & 14  & 0.64         & 0.64     \\
Duration                 & 18  & 0.78         & 0.89     \\
Finish the sentence      & 10  & 0.9          & 0.8      \\ \bottomrule
\end{tabular}
}
    \caption{Accuracy by reasoning types. $^*$ denotes statistically significant improvement with $p$-val < 0.005.}
    \label{tab:acc_by_reasoning}
\end{table}

\paragraph{\compact\ is significantly slower than zero-shot prompting:} The improved performance of \compact\ comes at the cost of longer running time and cost. With an average of 30 examples, \compact\ needs to handle 4.4 times more tokens in the prompt and generate 1.3 times more tokens owing to the intermediate steps.

\paragraph{Specific examples where \compact\ helps:} To better understand \compact, we qualitatively analyze 40 examples for which zero-shot GPT-4 generates incorrect answers while \compact\ answers correctly. This analysis reveals two key advantages of \compact. First, while zero-shot prompting is reasonably good at finding salient information from the input document, its generative answers tend to be based only on \emph{local} context around this information. For instance, when asked about the number of wives the character ``Dan Merrol'' has, the baseline successfully identifies six names that appear to be Dan's wives. However, \compact\ takes into account the revelation that these names ``\textit{were actually memories from the brain donors whose parts were used to reconstruct his brain}'' and thus correctly reasons that Dan only has one wife. In this case, \compact\ provides answer that demonstrates a more comprehensive understanding of the entire article. Second, \compact\ generates more detailed and thorough answers. For instance, given the question \emph{``Why is Kumaon a good region for potential forest preservation?''}, the zero-shot answer considers only one aspect of the reason, whereas \compact\ elaborates on multiple aspects. This allows \compact's answer to be mapped to the correct option (``All other choices''), while the zero-shot answer maps to the option corresponding to the single aspect.

\paragraph{Where does \compact\ go wrong?} We additionally examine 40 examples for which \compact\ answers incorrectly, and we categorize the errors into three categories (detailed examples and explanations in Table~\ref{tab:err_examples}):
\begin{itemize}
    \item \textbf{True negatives:} Questions for which \compact's generative answer is mapped to the wrong option. This category can be further divided into two subcategories: (1) cases where the plan has critical issues, and (2) cases where the plan is satisfactory but the intermediate execution produces incorrect output. Out of the 40 examples, 29 are true negatives, with 7 plan errors and 22 execution errors.
    \item \textbf{False negatives:} Questions for which \compact's generative answers are correct but incorrectly mapped to the wrong option. This kind of error is unavoidable as we use LLM for automatic answer mapping. Out of the 40 examples, 5 are false negatives. 
    \item \textbf{Other:} Some QuALITY questions are heavily dependent on the options; that is, the correct answer can only be determined after examining all the options. For instance, Table~\ref{tab:err_examples} presents a question asking who would enjoy the story the most of the given options. Although \compact\ offers an answer based on the story's genre---which is not incorrect---it is not as accurate as the gold label. Furthermore, there are instances where the model's free-form answers lack sufficient details and can thus be mapped to more than one option or no options at all. We classify these responses as a separate category. Out of 40 examples, 6 fall into this \textbf{Other} category.
\end{itemize}

\begin{table}[]
    \centering
    \begin{tabular}{lr}
    \toprule
        \textbf{Human annot. category }& \textbf{\# of plans} \\\midrule
        Unnecessary steps &  15 \\
        Steps can be merged & 2 \\
        Plan misses information & 3 \\
        Plan may lead to incorrect answer & 4 \\
        Plan needs slight edit & 7 \\
    \bottomrule
    \end{tabular}
    \caption{human freeform feedback aggregation}
    \label{tab:human_feedback}
\end{table}

\paragraph{Human evaluation of model-generated plans:}  \label{sec:human_eval}
The quality of plans generated by \compact\ is critical, as they serve as the basis for the plan execution stage. To gain further insight on the quality of these plans, we perform a human evaluation by hiring annotators on Upwork\footnote{We pay the annotators at the rate of \$25/h.} to provide feedback on the generated plans.\footnote{We provide a few examples in Appendix~\ref{sec:human_eval}.} 
Concretely, we ask annotators to assess (1) the correctness of the plans (binary choice), assuming error-free execution at each step, and (2) provide free-form feedback on any flaws or potential improvements. On average, annotators regard over 97\% of all plans as correct, with over 94\% confidence, although these numbers are inflated because the annotators do not have access to the long story when making these judgments. More interestingly, Table~\ref{tab:human_feedback} displays their feedback aggregated over common themes, which shows that the primary issue with existing plans is the presence of unnecessary steps (10\% of the total annotated plans). 
Annotators also notice that GPT-4 can be inattentive to subtle details while generating plans.
For example, given the question ``\emph{Do you think it would be fun to live in the universe in which this story takes place?}'', the model decides to ``\emph{evaluate the pros and cons of living in the universe based on the features found in the input article}''. However, human annotator argues that ``\emph{just because something is positive doesn't necessarily mean it is ``fun''. Any pros on the list might outweigh the dangers noted, resulting in an incorrect answer of 'yes'...}".

\section{Conclusion}
In this work, we introduce \compact, a framework for tackling complex reasoning over long documents. To answer a question, \compact\ first proposes a plan based on a set of actions mined from a training set, and then it executes the plan step by step via prompting itself with a template filled with output from previous stages. We demonstrate the effectiveness of \compact\ on a challenging subset of QuALITY. Experiments and analysis show that prompting GPT-4 with \compact\ yields more accurate and comprehensive answers than zero-shot and chain-of-thought prompting, and human annotators judge the generated plans to be reasonable.
\section*{Limitations}
While \compact\ shows promising results for long document reasoning, there are several limitations to our approach. Like other prompting methods, \compact\ is susceptible to generating misinformation or hallucinations. It is also more time-consuming and computationally costly than the baseline approach of directly prompting an LLM to answer the question. Moreover, \compact\ may over-complicate simple questions that only need superficial reasoning over long-form narratives. Finally, \compact\ is still bounded by the maximum context window size of the LLMs. Overall, our work leaves many interesting directions in this space (e.g., new datasets, modules, stage refinements) open for exploration.



\bibliography{anthology,custom}
\bibliographystyle{acl_natbib}

\appendix

\section{GPT-4 Multiple-choice setup performance} \label{sec:multi_choice_setup}
While our primary focus is on the generative QA setup in the main text, we provide GPT-4's performance under the standard multiple-choice setup here in the Appendix. On the entire QuALITY dev set, GPT-4 achieves an accuracy of 84.4\%. For the 1000 challenging question set, GPT-4 reaches an accuracy of 78.7\%, nearly 10 points higher than the GPT-4 zero-shot generative baseline. This result suggests that there is still room for improvement in GPT-4's generative answers. We also observe that GPT-4 is sensitive to the ordering of the provided options. We further evaluate GPT-4 with three shuffled versions of the options (swap A and D, B and C; swap A and C, B and D; swap A and B, C and D). While the overall accuracy of these versions remains similar, the questions that are consistently answered correctly across all four option orderings drop to 68.7\%. This result raises the question of whether GPT-4 truly ``understands'' the question and further motivates the generative QA setup.

\section{Verify Accuracy of Answer Mapping} \label{sec:human_answer_map}
As demonstrated in Section~\ref{sec:analysis}, the mapping stage is not always reliable. To understand the frequency of mapping errors, we conduct a small-scale human answer mapping study. We recruit three professionals on Upwork. We randomly select 50 questions and ask annotators to read \compact\ output and then map it to one of the provided options. On average, annotators agree with $\sim$83\% of GPT-4 mappings, with inter-annotator agreement on four-class settings of $\kappa$ = 0.677. For questions where annotators disagree with each other or do not concur with GPT-4, they tend to be those that can be mapped to than one option or none of the options. We believe this level of accuracy is decent enough to let GPT-4 perform the mapping step for evaluation.

\section{Can \compact\ benefit from more human-written examples?}
While we have employed self-refinement and executed the model-generated plan to ensure the quality of ICL demonstrations, it is natural to ask if we can further improve \compact\ by incorporating more quality-assured human-written examples. Therefore, we evaluate an alternative version of \compact\ in which the in-context examples for plan generation are replaced with 11 human-written examples. This variant achieves 70.3, 76.8, and 72.3 on the long split, the short split, and the total evaluation data, respectively. These results suggest that additional human input may note be necessary to achieve strong results.

\section{Prompts and templates used in \compact} \label{sec:prompts}

\begin{table*}[]
    \centering
    \footnotesize
\scalebox{1.0}{
    \begin{tabular}{p{\linewidth}}
    \toprule
       \textbf{Prompt for Action Mining}   \\ \midrule
       {[}Actions{]}\\ - CONCAT(S1, S2, ...) : Concatenate the input S1, S2, ...\\ - EXTRACT(CTX, X) : Extract the exact wording that X is referring to from input CTX.\\ - FIND\_X(CTX, X): Find and summarize all relevant information about X in the input CTX. \\ - FIND\_REASON(CTX, X) : Find and summarize the cause or reason of X given input CTX.\\ - FIND\_MORAL(CTX) : Find the intended lesson or moral of the input CTX.\\ - SUMMARIZE(CTX): Provides a general summary about the given CTX.\\ - SUMMARIZE\_X(CTX, X) : Provides a summary about X given the provided input CTX.\\ \\ \\ {[}Instructions{]}\\ Suppose you are given a question about an article as well as a list of actions that you can execute to solve the question (shown above). You can imagine the actions as functions in a program, where you have input arguments and output. The output of an action can be fed as input to another action. The output of the final action will be the answer to the given question. Suppose you haven't read the article yet, please present a sequence of actions that you would use to answer the question. \\ \\ Here are a few examples:\\ \\ Question:\\ What is the “space cafard” that Si describes?\\ \\ My new actions:\\ - COMPREHEND(CTX, X) : Provide a detailed comprehension of X given the input CTX.\\ \\ My sequence of actions:\\ 1. snippet = EXTRACT(CTX, "space cafard") : Extract the exact wording regarding "space cafard" from the input CTX.\\ 2. ans = COMPREHEND(CTX, X) : Provide a detailed comprehension of the input X given the input CTX.\\ \\ \\ \\ Question:\\ Why did the author write the article?\\ \\ My new actions:\\ - None\\ \\ My sequence of actions:\\ 1. moral = FIND\_MORAL(CTX) : Find the intended lesson or moral of the input CTX.\\ \\ \\ Your answer must follow the following rules: 1. The present sequence should be minimal, i.e., no unnecessary actions.  2. The sequence of actions should be specific and cover every detail about the question.  3. The sequence of actions should use as many as existing actions as possible. 4. It is fine to create new actions, however, the created new actions should be maximally reusable and generalizable to other reading comprehension questions.  5. The arguments should cover all the details of the given question.\\ \\ {[}Question{]}\\ \textcolor{purple}{\{Question\}}\\ \\ {[}Answer{]}\\ Now please provide the plan for the above question.\\ Your answer should follow the format: \\ \\ My new actions (if any):\\ - my\_new\_action\_1(here goes the arguments) : {[}one-sentence explanation{]}\\ - my\_new\_action\_2(here goes the arguments) : {[}one-sentence explanation{]}\\ ...\\ \\ My sequence of actions:\\ 1. output\_1 = action\_1(here goes the arguments) : {[}one-sentence explanation{]}\\ 2. output\_2 = action\_2(here goes the arguments) : {[}one-sentence explanation{]}\\ ...\\ 
       \\
    \bottomrule
    \end{tabular}}
    \caption{Prompt for action mining. \textcolor{purple}{\{Question\}} indicates the placeholder for filling in training set question. In this stage, we only care about the new actions proposed by the model.}
    \label{tab:prompt_action_mining}
\end{table*}
\begin{table*}[]
    \centering
    \footnotesize
    \scalebox{0.9}{
    \begin{tabular}{p{\linewidth}}
    \toprule
       \textbf{Mined Actions after reducing number of actions with LLM}   \\ \midrule
ANALYZE(CTX, X, Y) \# Analyze the relationship, attitude, or feelings between X and Y, or the character, language, tone, or symbolism of X given the input CTX. \\ COMPARE(CTX, X, Y, Z) \# Compare X and Y in the context of Z, considering aspects such as abilities, assets, attractiveness, behavior, concerns, contributions, cultures, events, experiences, feelings, ... \\COMPREHEND(CTX, X) \# Provide a detailed comprehension of X given the input CTX. \\CONCAT(S1, S2, ...) \\DEFINE(CTX, X) \# Provide the definition of X given the input CTX. \\DESCRIBE(CTX, X, Y) \# Provide a description of X in terms of Y, such as character, genre, or introduction given the input CTX. \\EVALUATE(CTX, X, Y) \# Evaluate aspects such as feeling, outcome, performance, personalities, risk, or truth of X in relation to Y given the input CTX. \\EXCEPT(CTX, LIST) \# Find the item that is not mentioned in the input CTX but is present in the given..\\EXPLAIN\_PROCESS(CTX, X) \# Provide a detailed explanation of the process X given the input CTX. \\FIND\_BARRIERS\_CAUSES(CTX, X) \# Find and summarize the remaining barriers or causes related to X given the input CTX. \\FIND\_BEHAVIOR\_REASON(CTX, X) \# Find the reason behind the behavior X given the input CTX. \\FIND\_BENEFIT(CTX, X) \# Find the direct benefit of X given the input CTX. \\FIND\_BEST(CTX, X, Y) \# Find the best X in the context of Y given the input CTX. \\FIND\_CHARACTER(CTX, X) \# Find and summarize the character traits, transformation, and changes of X given the input CTX. \\FIND\_COMMON(CTX, X, Y, Z) \# Find the common ground, characteristics, or commonalities between X, Y, and Z given the input CTX. \\FIND\_CONDITION(CTX, X, Y) \# Find the condition, outcome, or consequences related to X and Y given the input CTX. \\FIND\_CONFLICT\_CONCERN(CTX, X, Y) \# Find the conflict, concern, or disagreement between X and Y given the input CTX. \\FIND\_CONSISTENCY(CTX, X) \# Determine if X is consistent throughout the input CTX. \\FIND\_DECISION(CTX, X) \# Find the decision, factor, or event that influenced X's decision in the input CTX. \\FIND\_DESCRIPTION(CTX, X) \# Find all descriptions, characteristics, or words that describe X given the input CTX. \\FIND\_DETAILS(CTX) \# Find all the details about a topic (e.g., contract, city-state) discussed in the input CTX. \\FIND\_DIALOGUE(CTX, X, Y) \# Find the dialogue between X and Y in the input CTX. \\FIND\_DIFFICULTY\_DANGER(CTX, X) \# Find the most difficult aspect, challenge, or danger faced by X in the given input CTX. \\FIND\_ELEMENT(CTX, X, Y) \# Find the element X related to Y given the input CTX. This function can cover message, method, metrics, mismatch, mission, mistake, most likely, motif, motivation, nationalities, negative critique, negative effect, next event, normal, objective, obstacles, ... \\FIND\_EMOTION(CTX, X, Y) \# Find the emotion or feeling X feels towards Y given the input CTX. \\FIND\_ENDING(CTX, X) \# Find the ending or conclusion of X's story or the input CTX. \\FIND\_EVENT(CTX, X) \# Find the event involving X in the input CTX (e.g., betrayal, change, climax). \\FIND\_EVIDENCE\_EXAMPLE(CTX, X) \# Find evidence or an example supporting X given the input CTX. \\FIND\_EXCEPTION(CTX, X, Y, Z) \# Find the exception or characteristic that is not common among X, Y, and Z given the input CTX. \\FIND\_EXPECTATION(CTX, X) \# Find the expectation, assumption, or impact about X given the input CTX. \\FIND\_EXPLANATION(CTX, X) \# Find the most likely explanation, critique, or doubt for X given the input CTX. \\FIND\_FACT\_FALSE(CTX, X) \# Find a definite fact or false statement about X given the input CTX. \\FIND\_FEARS\_DISTRACTIONS(CTX, X) \# Find the fears, concerns, or distractions of X given the input CTX. \\FIND\_FEATURES(CTX, X) \# Find all the features that X cares about given the input CTX. \\FIND\_FIRST\_INSTANCE(CTX, X) \# Find the first instance of X happening in the input CTX. \\FIND\_FLAW(CTX, X) \# Find the greatest flaw of X given the input CTX. \\FIND\_FOCUS(CTX, X) \# Find the person or object that is focused on the most in the input CTX, given a list of X. \\FIND\_FORESHADOW(CTX, X, Y) \# Find the instance where X foreshadows Y in the input CTX. \\FIND\_FUTURE(CTX, X) \# Find the future, predicted outcome, or action of X given the input CTX. \\FIND\_GRIEVANCE(CTX, X) \# Find and summarize the grievance X has against something or someone in the input CTX. \\FIND\_HALO\_EFFECT(CTX, X) \# Find and summarize one halo effect of X given the input CTX. \\FIND\_HUMBLENESS(CTX, X) \# Find the instances of humbleness presented by X in the input CTX. \\FIND\_HYPOTHETICAL(CTX, X) \# Find the hypothetical outcome or consequence of X given input CTX. \\FIND\_IMAGINATION(CTX, X) \# Find and summarize how X imagines something in the input CTX. \\FIND\_IMPACT(CTX, X, Y) \# Find the event or experience that had the strongest impact on X's Y given the input CTX. \\
...\\
    \bottomrule
    \end{tabular}}
    \caption{A subset of mined actions from training set questions.}
    \label{tab:mined_actions}
\end{table*}

\begin{table*}[]
    \centering
    \footnotesize
    \scalebox{0.9}{
    \begin{tabular}{p{\linewidth}}
    \toprule
       \textbf{Prompt for Generating Plan}   \\ \midrule
{[}Actions{]}\\ ANALYZE(CTX, X, Y) \# Analyze the relationship, attitude, or feelings between X and Y, or the character, language, tone, or symbolism of X given the input CTX.\\ COMPARE(CTX, X, Y, Z) \# Compare X and Y in the context of Z, considering aspects such as abilities, assets, attractiveness, behavior, concerns, contributions, cultures, events, experiences, feelings, focus, intelligence, irony, nationalities, performance, praise, reactions, reviews, secretiveness, time periods, treatment, truth, or worlds given the input CTX.\\ COMPREHEND(CTX, X) \# Provide a detailed comprehension of X given the input CTX.\\ CONCAT(S1, S2, ...)\\ DEFINE(CTX, X) \# Provide the definition of X given the input CTX.\\DESCRIBE(CTX, X, Y) \# Provide a description of X in terms of Y, such as character, genre, or introduction given the input CTX.\\ EVALUATE(CTX, X, Y) \# Evaluate aspects such as feeling, outcome, performance, personalities, risk, or truth of X in relation to Y given the input CTX.\\ ...\\ \{\textcolor{blue}{List of Actions as shown in Table}~\ref{tab:mined_actions}\}\\ \\ {[}Instructions{]}\\ Suppose you are given a question about an article, as well as a list of potential actions (shown above) that you can execute to solve the question . You can imagine the actions as functions in a program, where you have input arguments and output. The output of an action can be fed as input to another action. Please present a sequence of actions that you would use to answer the question after you read the article. The sequence of actions should be specific and cover all the details about the question. Please prioritize using the actions presented in the list above. If you need to add new actions, please follow the format below. Please assign the output of each action with a distinct name, which can be passed into other actions as argument. Think twice before you provide your answer. Make sure your answer is valid, clear, and easy to understand. Keep the answer simple and remove any unnecessary steps. Do not use list comprehension or dictionary comprehension. Keep each action minimally simple. If a question is unanswerable (e.g., requires options), collect as much information as possible from the input such that it will be answerable when provided with options. Your answer should follow the format: \\ '''\\ New actions:\\ - new\_action\_1(arguments) : {[}one-sentence general explanation{]} or "-None" if there no need to add new actions\\ - new\_action\_2(arguments) : {[}one-sentence general explanation{]} or "-None" if there no need to add new actions\\ \\ 1. output\_1 = action\_1(here goes arguments) : {[}one-sentence explanation{]}\\ 2. output\_2 = action\_2(here goes arguments) : {[}one-sentence explanation{]}\\ ...\\ '''\\ \\ The following are a few examples\\ \\ Question: "How do Ross and Mehta view Brown's acquisition of the magazine?"\\ \\ Answer:\\ New actions:\\ - FIND\_OPINION(CTX, X, Y) : Find the opinion of X about Y given the input CTX\\ \\ 1. ross = FIND\_CHARACTER(CTX, "Ross") : Identify who Ross is in the input article\\ 2. mehta = FIND\_CHARACTER(CTX, "Mehta") : Identify who Mehta is in the input article\\ 3. brown = FIND\_CHARACTER(CTX, "Brown") : Identify who Brown is in the input article\\ 4. magazine\_acquisition = FIND\_EVENT(CTX, "Brown's acquisition of the magazine") : Find the event of Brown's acquisition of the magazine in the input article\\ 5. ross\_opinion = FIND\_OPINION(CTX, ross, magazine\_acquisition) : Find the opinion of Ross about Brown's acquisition of the magazine\\ 6. mehta\_opinion = FIND\_OPINION(CTX, mehta, magazine\_acquisition) : Find the opinion of Mehta about Brown's acquisition of the magazine\\ 7. ans = CONCAT(ross\_opinion, mehta\_opinion) : Combine the opinions of Ross and Mehta on Brown's acquisition of the magazine to form the final answer\\ ... \{\textcolor{gray}{more few-shot examples}\} ...\\ \\ \\ {[}Question{]}\\ Now you are given a question about an article:\\     \textcolor{purple}{\{question\}}\\ Please provide a plan (sequence of actions) that can arrive to the answer after reading the article. As the corresponding options are not provided for the question, when the question is not answerable without the options, simply collect as much information as possible from the input such that it will be answerable with the options. Make sure the plan you generate is valid and faithful to the question.\\ \\ \\ {[}Answer{]} \\
    \bottomrule
    \end{tabular}}
    \caption{Prompt for generating plan given a question, which is filled in the placeholder \textcolor{purple}{\{question\}}.}
    \label{tab:prompt_plan_generation}
\end{table*}
\begin{table*}[]
    \centering
    \footnotesize
    \begin{tabular}{p{\linewidth}}
    \toprule
       \textbf{Prompt for Executing Single Step of the Plan}   \\ \midrule
       Article\\
       \textcolor{purple}{\{Long document\}} \\
       End of Article\\
       ---\\
       Please read the above text first, and then follow the instructions below.\\ \\
       
    {[}Instructions{]}\\
\\
\textcolor{purple}{ \{Mined action and corresponding definition of current step. Example shown below.\} }\\
\textcolor{gray}{FIND\_EMOTION(CTX, X, Y) \# Find the emotion or feeling X feels towards Y given the input CTX.}\\
\\
\textcolor{purple}{ \{Current step in the plan generated in the previous stage. Example shown below.\}}\\
\textcolor{gray}{kolin\_opinion = FIND\_EMOTION(CTX, kolin, ``becoming a tree'')}\\
\\

\textcolor{purple}{\{Value assignment of input argument(s)\}}\\
\textcolor{gray}{X = ``In the story, Kolin is a steward from the Planetary State of Haurtoz who is part of a scouting party sent to explore a planet after their ship, the Peace State, is damaged. Kolin is unhappy with the oppressive regime on Haurtoz and dreams of escaping it. While exploring the planet, he encounters a tree named Ashlew, which is actually a man who has transformed into a tree. Ashlew tells Kolin about the Life, a powerful entity on the planet that can help individuals change their form...\{Output from previous step.\}''} \\
\textcolor{gray}{Y = ``becoming a tree''}
 \\ \\

{[}Answer{]}\\
\textcolor{purple}{\{A brief description of current step.\}}\\
\textcolor{gray}{(Find the emotion or feeling Kolin has towards becoming a tree himself in the input article)}\\
    \bottomrule
    \end{tabular}
    \caption{Prompt for executing a step in a plan. Prompt of this step is a template with placeholders which will be filled with the output from previous step(s). }
    \label{tab:prompt_plan_execution}
\end{table*}
\begin{table*}[]
    \centering
    \footnotesize
    \begin{tabular}{p{\linewidth}}
    \toprule
       \textbf{Prompt for Answer Mapping}   \\ \midrule
Relevant information for answering the question:\\
\\
\textcolor{purple}{\{Generative answer output by \compact.\}}\\ \\
Question: \textcolor{purple}{\{question\}} \\
\textcolor{purple}{\{Options\}}\\
\\
\\
Read the relevant information about the article and answer the question by selecting the best option above. Only one of them is correct.\\ \\ 
Answer (select from A, B, C, D):\\
    \bottomrule
    \end{tabular}
    \caption{Prompt for mapping generative answer to provided options}
    \label{tab:prompt_answer_mapping}
\end{table*}

\section{Human feedbacks on model-generated plan} \label{sec:human_eval}
\begin{table*}[]
    \centering
    \footnotesize
    \scalebox{1.0}{
    \begin{tabular}{p{0.3\linewidth} p{0.35\linewidth}p{0.25\linewidth}}
\toprule
\textbf{Question \& Options} & \textbf{Plan} & \textbf{Human feedback} \\
\midrule

What was Mrs. Jamieson's biggest problem in the story?

A: She did not understand the Stinson Effect

B: She had to raise her son alone

C: She was just able to make ends meet

D: She had to hide her scar

&

1. mrs\_jamieson = IDENTIFY\_ELEMENT(CTX, "Mrs. Jamieson") : Identify who Mrs. Jamieson is in the input article

2. mrs\_jamieson\_problems = FIND\_ELEMENT (CTX, "problems", mrs\_jamieson) : Find and list all the problems faced by Mrs. Jamieson in the story

3. biggest\_problem = FIND\_BEST(CTX, "problem", mrs\_jamieson\_problems) : Find the biggest problem faced by Mrs. Jamieson in the story

4. ans = CONCAT(mrs\_jamieson, biggest \_problem) : Combine Mrs. Jamieson and her biggest problem to form the final answer
\vspace{0.3em}

& 

Steps 2 \& 3 can be combined into one. mrs\_jamieson\_biggest

\_problems = FIND\_ELEMENT (CTX, "biggest problem", mrs\_jamieson): Finds the biggest problem faced by Mrs. Jamieson in the story \\ \hdashline 

\vspace{0.1em}Which class was the least intense?

A: Kung Fu

B: Tai Chi

C: Aikido

D: Brazilian Jujitsu

&

\vspace{0.1em}1. classes = FIND\_DETAILS(CTX) : Find all the details about the classes and their intensity levels mentioned in the input article

2. least\_intense\_class = FIND\_LEAST\_DESCRIBING\_WORD (classes, ``intense'') : Find the class that is least intense in the given input CTX

3. ans = CONCAT (least\_intense\_class, classes) : Combine the least intense class and the intensity levels of all classes to form the final answer \vspace{0.3em}

&

\vspace{0.1em}Adding the details for the other classes is unnecessary since the question is looking for a single answer, the least intensive class. \\ \hdashline

\vspace{0.1em}Do the Terrans ever come close to winning the battle within the story?

A: No, they continually lose

B: They win the whole battle with less casualties

C: Yes, by the surprise squadron Evelyn leads

D: Yes, by Evelyn cloning soldiers into battle

&

\vspace{0.1em}1. terrans = IDENTIFY\_ELEMENT(CTX, "Terrans") : Identify who the Terrans are in the input article

2. battle = FIND\_EVENT(CTX, "battle") : Find and summarize the battle event within the story in the input article

3. terrans\_battle = FIND\_RELATION(CTX, terrans, battle) : Find and summarize the Terrans' involvement in the battle from the input article

4. terrans\_close\_win = FIND\_CONDITION (CTX, "Terrans", "close to winning") : Find the condition or events where the Terrans come close to winning the battle in the input article

5. ans = CONCAT(terrans\_battle, terrans\_close\_win) : Combine the Terrans' involvement in the battle and the events where they come close to winning to form the final answer\vspace{0.3em}

&

\vspace{0.1em}Step 2 and 3 can be combined: Find and summarize the Terrans' battle event within the story in the input article \\ \hdashline

\vspace{0.1em}What level of depth does the author provide on the subjects they use to make their case?

A: Language is really the only thing covered in any depth

B: A broad, but not very deep assessment

C: They provide the reader with deeper arguments about the monetary system and striking tendencies than anything else

D: They provide deep, explanatory statistics to most arguments

&

\vspace{0.1em}1. author = IDENTIFY\_ELEMENT(CTX, "author") : Identify the author of the article

2. subjects = FIND\_ELEMENT(CTX, "subjects", author) : Find and list all the subjects the author uses to make their case in the input article

3. depth\_analysis = ANALYZE(CTX, subjects, author) : Analyze the level of depth the author provides on the subjects they use to make their case in the input article

4. ans = CONCAT(subjects, depth\_analysis) : Combine the subjects and the depth analysis to form the final answer for comparing with the options

&

\vspace{0.1em}Very good plan. \\

\bottomrule
\end{tabular}}
    \caption{Example human feedback from annotators on \compact-generated plans.}
    \label{tab:err_examples}
\end{table*}

\end{document}